\documentclass[runningheads]{llncs}
\usepackage[accepted]{llncsconf}
\conference{BNAIC/BeneLearn 2020}
\usepackage{graphicx}
\usepackage{amsmath,amssymb,amsfonts}
\usepackage[font=small,labelfont=bf]{caption}
\usepackage{tikz}
\usetikzlibrary{shapes.geometric,arrows.meta,positioning,decorations.markings,decorations.pathreplacing}
\tikzstyle{my right of} = [right=of #1.east]
\tikzstyle{my left of} = [left=of #1.west]
\usepackage{subcaption}
\usepackage{multirow}
\usepackage{multicol}
\usepackage{array}
\usepackage[export]{adjustbox}
\usepackage{comment}

\begin{document}
\title{Does the dataset meet your expectations? Explaining sample representation in image data\thanks{Work supported by the Wallenberg Artificial Intelligence, Autonomous Systems and Software Program (WASP), funded by the Knut and Alice Wallenberg Foundation.\looseness=-1}}
\titlerunning{Does the data meet your expectations?}
%
\author{Dhasarathy Parthasarathy\inst{1, 2} \and
Anton Johansson\inst{2}}
\authorrunning{Parthasarathy and Johansson}
%
\institute{Volvo Group, Sweden \\
\email{dhasarathy.parthasarathy@volvo.com}\\\and
Chalmers University of Technology, Sweden\\
\email{johaant@chalmers.se}}
\maketitle              
\begin{abstract}
Since the behavior of a neural network model is adversely affected by a lack of diversity in training data, we present a method that identifies and explains such deficiencies. When a dataset is labeled, we note that annotations alone are capable of providing a human interpretable summary of sample diversity. This allows explaining any lack of diversity as the mismatch found when comparing the \textit{actual} distribution of annotations in the dataset with an \textit{expected} distribution of annotations, specified manually to capture essential label diversity. While, in many practical cases, labeling (samples $\rightarrow$ annotations) is expensive, its inverse, simulation (annotations $\rightarrow$ samples) can be cheaper. By mapping the expected distribution of annotations into test samples using parametric simulation, we present a method that explains sample representation using the mismatch in diversity between simulated and collected data. We then apply the method to examine a dataset of geometric shapes to qualitatively and quantitatively explain sample representation in terms of comprehensible aspects such as size, position, and pixel brightness.\looseness=-1

\keywords{Sample selection bias \and Explainability \and Outlier detection.}
\end{abstract}

\section{Introduction}
Choosing the right data has always been an important precondition to deep learning. However, with increasing application of trained models in systems which are required to be dependable (\cite{DBLP:journals/corr/abs-1912-10773}, \cite{DBLP:journals/information/BermanBCC19}), there is increasing emphasis on making this choice well-informed (\cite{DBLP:journals/corr/abs-1812-05389}, \cite{DBLP:conf/re/VogelsangB19}). Consider the perception system of a self-driving vehicle which is partially realized using deep learning and is expected to dependably detect pedestrians. To ensure that the system meets such an expectation, it is necessary to choose training and validation sets that adequately cover critical scenarios~(\cite{DBLP:journals/corr/SeshiaS16}, \cite{Thorn2018-09-01}) like residential areas and school zones, where the vehicle is likely to meet pedestrians. Choosing, conversely, a dataset that contains only scenes of motorway traffic, which does not cover many scenarios involving pedestrians, is likely to produce a trained model that violates expectations on pedestrian detection. Scenarios covered by a dataset may be considered sufficient when samples of adequate variety are represented in it. With practical image datasets typically being high-dimensional and large, posing and evaluating explicit conditions on the adequacy of sample representation is not straightforward. \looseness=-1
\vspace{-15pt}

\subsubsection{Interpretable assessment of sample representation}
Consider a traffic dataset $\mathcal{S}$ of images $X_i \sim P(X|Y)$ and annotations ${Y_i\sim P(Y)}$. A major practical concern in such datasets is whether it adequately represents corner cases like intersections with stop signs, roundabouts with five exits, etc. With the true/target distribution of traffic scenes $P(X,Y)$ clearly containing instances of such cases, \textit{any under-representation} in $\mathcal{S}$ can be broadly framed as shortcomings in data collection and processing, otherwise known as \textit{sample selection bias}(\cite{DBLP:conf/icml/Zadrozny04}). Given that the dataset is eventually used to train a model that is deployed in a safety-critical system, engineers may actively seek to properly comprehend and account for such bias. But how does one express such bias in human interpretable terms? One clue comes from annotations ${Y_i\sim P(Y)}$. In typical traffic datasets, $Y$ encodes object class labels and bounding box positions. If necessary and feasible, $Y$ can be expanded to contain information such as location, lighting conditions, weather conditions, etc. When $Y$ is adequately detailed, the distribution of annotations $P_{\mathcal{S}}(Y)$ clearly becomes a reasonable, low-dimensional, and therefore a human interpretable measure of sample representation in $\mathcal{S}$. Engineers can exploit this notion to \textit{specify} a distribution of annotations $P_\mathcal{T}(Y)$, expressing the sample representation that is \textit{expected} in the dataset. While the target distribution of annotations $P(Y)$ may be unknowable, $P_{\mathcal{T}}(Y)$ is an explicit declaration of the sub-space that the dataset is expected to cover at the minimum. If $\mathcal{S}$ is equivalently labeled, then selection bias (and thereby sample under-representation) is simply given by the mismatch between expectations $P_{\mathcal{T}}$ and reality $P_{\mathcal{S}}$. In practice, however, due to the effort and expense involved in labeling, $\mathcal{S}$ may either lack labels or may be completely unlabeled, meaning that $P_{\mathcal{S}}(Y)$ is often unavailable. Combining simulation, outlier detection, and input attribution, we show that it is possible to explain sample representation in a comprehensible low-dimensional form, even when annotations are not explicitly available in $\mathcal{S}$. \looseness=-1
\vspace{-8pt}

\subsubsection{Contributions} Delving into the less-explored area of \textit{explaining} sample representation in a dataset, we demonstrate a method that
\vspace{-2pt}
\begin{itemize}
    \item explains sample representation in interpretable terms for annotated data
    \item uses parametric simulation and outlier detection to do the same for non-annotated data
\end{itemize}
\vspace{-2pt}
In addition to visualization, we propose a quantitative explanation of sample under-representation using an \textit{overlap index}. Also, unlike existing methods that mainly address imbalances in available data, ours can explain gaps in the availability of data. Such an explanation helps engineers better understand data as a crucial ingredient of the training process. Downstream, this helps them re-asses data collection methods and to verify, reason, or argue about -- at times a requirement for standards compliance \cite{DBLP:conf/safecomp/BirchRHBBHJMP13} -- the overall dependability of the model trained with this data. Data and code used in this work are publicly available\footnote{https://github.com/dhas/SpecCheck}. \looseness=-1

\section{Explaining sample representation using annotations}
\label{sec:exp_bias_ann}
\subsubsection{Visualizing sample representation}
\begin{figure}[h]
    \centering
    \includegraphics[width=0.5\linewidth]{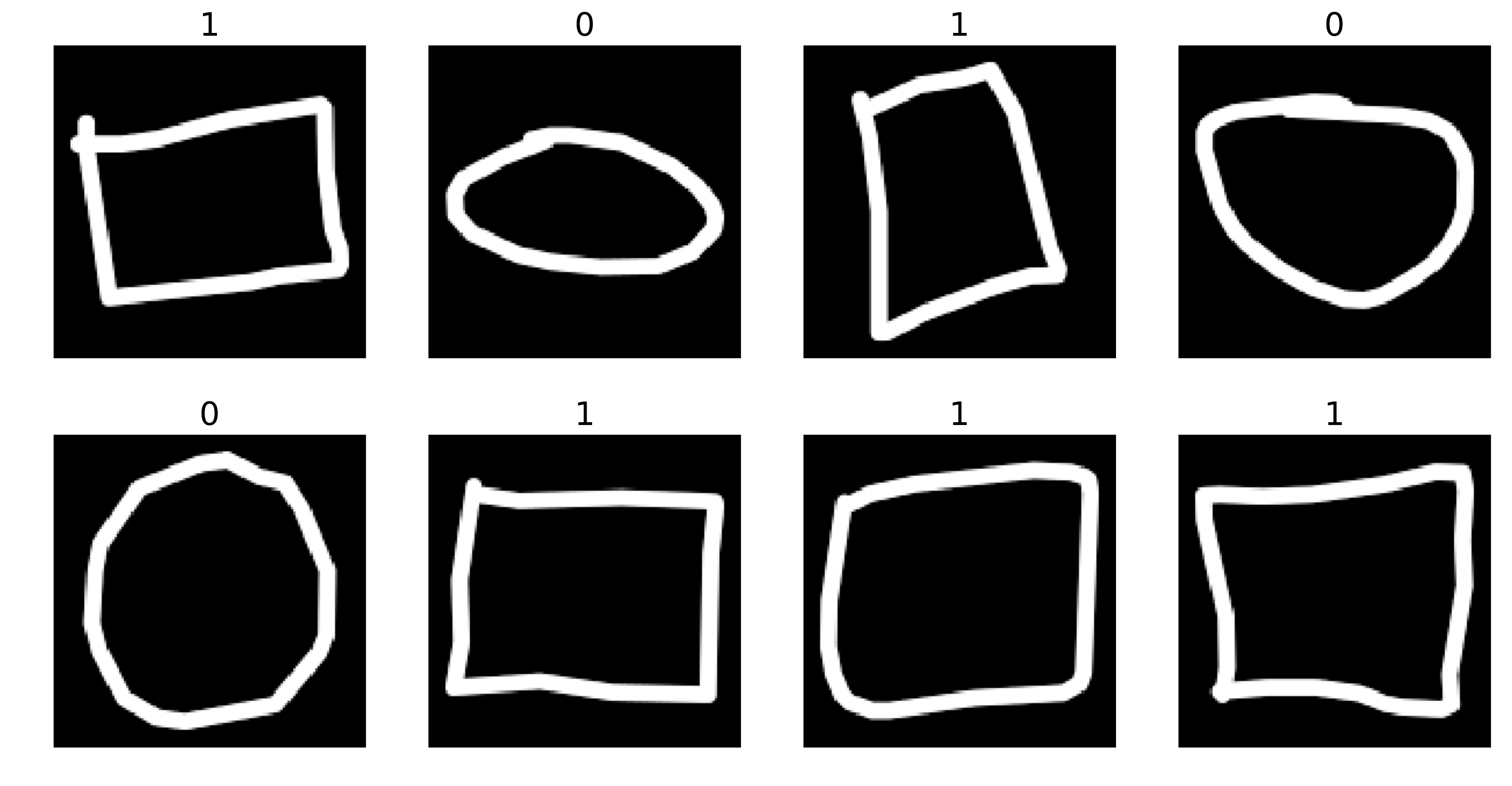}
    \setlength{\belowcaptionskip}{-15pt}
    \caption{Samples from the dataset $\mathcal{S}$. Only the class label $Y^1$ is available}
    \label{fig:qd_samples}
\end{figure}

We now introduce a simple running example of examining sample representation in a dataset $\mathcal{S}$ containing images of two hand-drawn shapes\footnote{Collected from Quick, Draw! with Google -- https://quickdraw.withgoogle.com/data} -- circles and squares  (Figure \ref{fig:qd_samples}). With the shape as the sole available label, one can define $\mathcal{S} = \{(X_i,Y^1_i)\},~i=1...N$, where $X_i$ is a grayscale image of size $(128,128)$ and $Y^1_i \in K=\{0,1\}$ is the shape label, corresponding to circle and square respectively. Understanding sample representation in this dataset may be necessary when it is a candidate for training a model that, for example, either recognizes or generates shapes. To ensure dependable model performance, system designers may want to confirm that images of adequate variety are represented in $\mathcal{S}$. In a dataset of grayscale geometric shapes, it is intuitive to analyze sample representation in terms of concerns such as the size and position of the shapes on the image canvas, and the average brightness of pixels in the shape. All these concerns can be captured by defining a 6-d annotation vector $Y = (Y^1,...,Y^6)$, including shape-type, which is known. With $\mathcal{U}$ denoting the discrete uniform distribution, designers can begin with defining an expected spread of shape-size using a latent label $Y^S \sim \mathcal{U}\{30, 120\}$, denoting the side-length in pixels of a square box bounding the shape. This can be followed by defining expectations on the spread of (i) the top-left corner of the bounding box, $Y^2, Y^3 \sim \mathcal{U}\{0,128-Y^S\}$, (ii) the bottom-right corner of the bounding box $Y^4, Y^5 \sim \mathcal{U}\{Y^S, 128\}$, and (iii) the average pixel brightness $Y^6 \sim \mathcal{U}\{100, 255\}$. Put simply, $P_{\mathcal{T}}(Y)$ expects shapes of a specified range of sizes and brightness to be uniformly represented in the dataset $\mathcal{S}$. All positions are also expected to be uniformly represented, as long as the shape can be fully fit in the image canvas.
\captionsetup[subfigure]{width=0.9\textwidth}

\begin{figure}[h]
  \begin{minipage}[t]{0.48\textwidth}
    \centering
    \includegraphics[width=\linewidth]{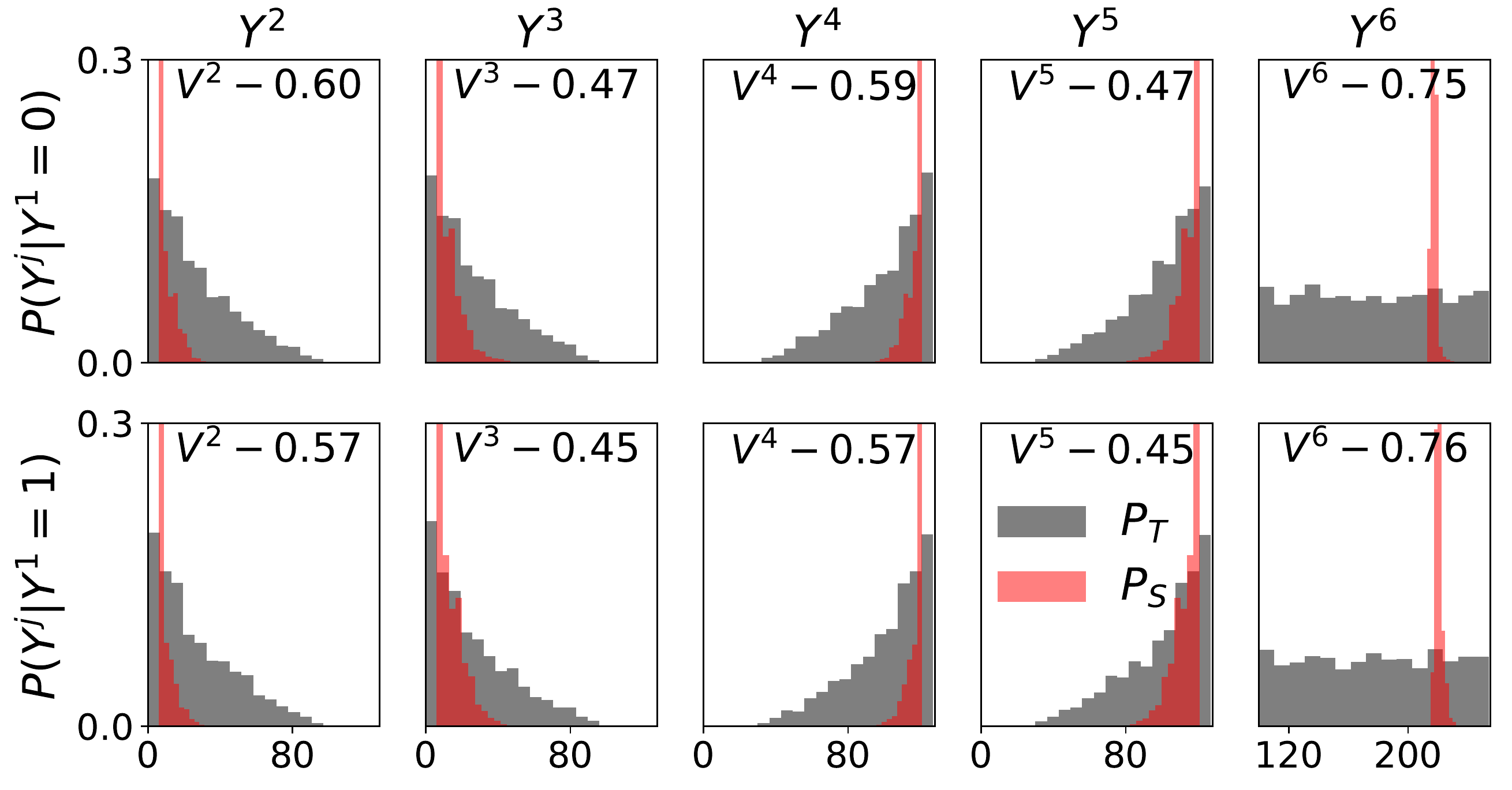}
    \captionof{figure}{Explaining sample representation}
    \label{fig:os_cs_ann}
  \end{minipage}
  \begin{minipage}[t]{0.48\textwidth}
    \centering
    \includegraphics[width=\linewidth]{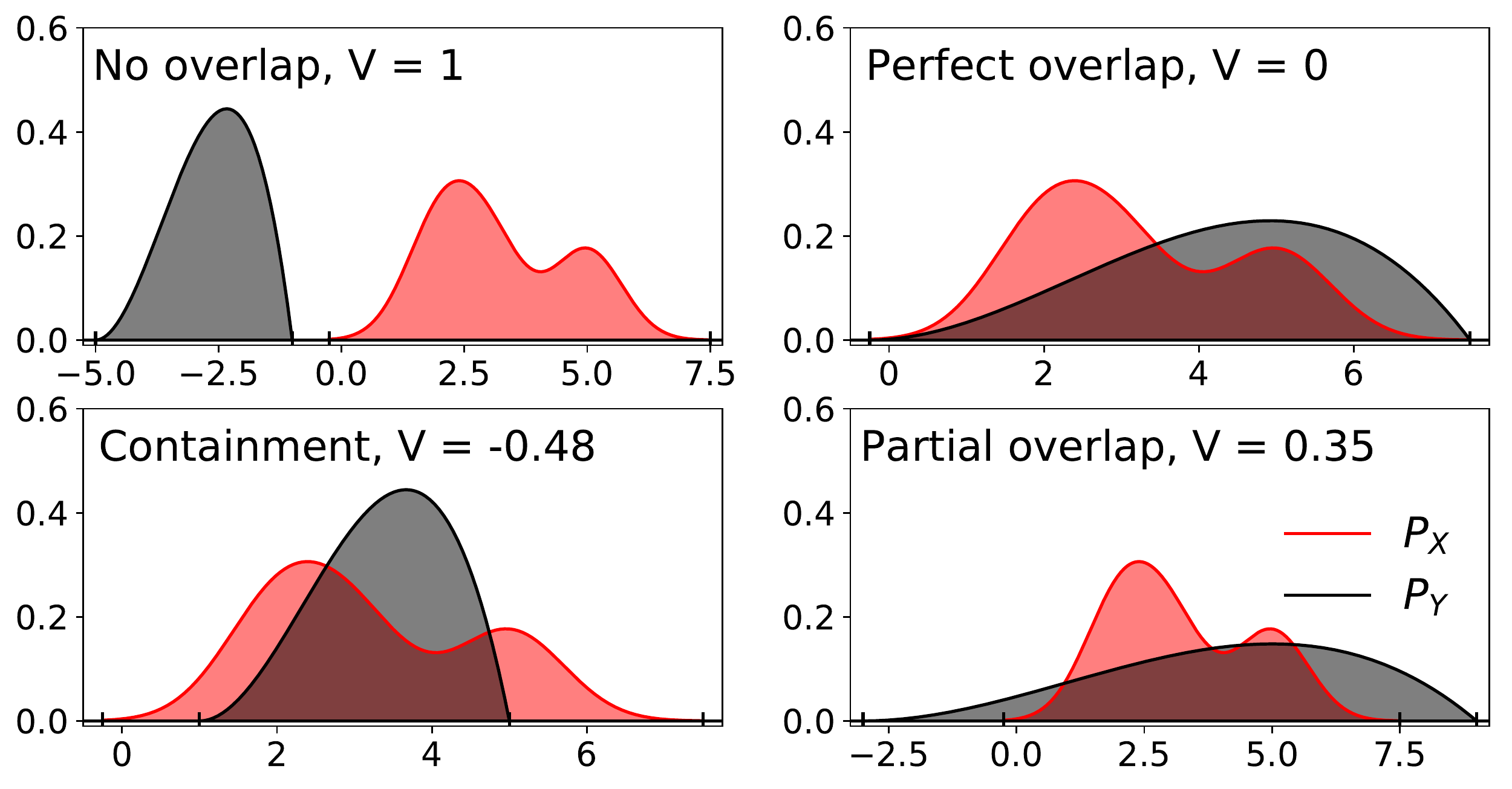}
    \captionof{figure}{Illustration of $V(P_X,P_Y)$}
    \label{fig:ovl_ind}
  \end{minipage}
\end{figure}

To illustrate the idea of explaining sample representation using annotations, an automatic labeling scheme $Y_i = L(X_i)$ is used to produce complete 6-d annotations for $X_i$. For circles and squares, it is easy to define a scheme that looks at the extent of the shape and draws bounding boxes. The average brightness is given by the mean of non-zero pixels in the canvas. The availability of labels $Y_i$ helps assemble the actual distribution of samples in the dataset $P_{\mathcal{S}}(Y)$, allowing direct comparison with expectations $P_\mathcal{T}(Y)$. Jointly visualizing label distributions for each shape (Figure \ref{fig:os_cs_ann}) shows that, along all design concerns $Y^j$, the spread of $P_{\mathcal{T}}$ (marked black) is much wider than the very narrow $P_{\mathcal{S}}$ (marked red). This shows that, while $P_{\mathcal{T}}$ expects shapes of a broad range of sizes, positions and brightness to be represented, $P_{\mathcal{S}}$ is clearly biased and massively over-represents large and bright shapes located in the center on the canvas. As long as the annotation vector $Y$ is of manageable length, joint visualization becomes an interpretable qualitative explanation of sample representation in the dataset. \looseness=-1

\vspace{-20pt}

\subsubsection{Quantifying sample representation}
By framing sample selection bias, and thereby sample under-representation, as the mismatch between expected and true label probability distributions, it becomes possible to quantify it using measures of statistical similarity. Choosing the right measure, however, requires a proper understanding of the nature of each distribution. Having calculated it using true labels of each sample, it is clear that $P_\mathcal{S}(Y)$ represents the actual sample distribution in $\mathcal{S}$. The distribution of expectations $P_\mathcal{T}$ is of a slightly different nature and, to better understand it, let us consider the expectation $P_\mathcal{T}(Y^6)= \mathcal{U}\{100, 255\}$, placed on the representation of average brightness of shapes in the dataset. While the expectation on brightness being spread between specified lower and upper limits is strict, imposing the spread to be uniform is arbitrary. This is a deliberate measure of simplification to ease the considerable burden in modeling expectations $P_\mathcal{T}$ and let it capture the critical range of interest in the target distribution. Put simply, expected sample representation is primarily encoded by the \textit{support} (\ref{eq:supp}) of $P_\mathcal{T}$. By specifying strict support, but arbitrary distribution of mass, sample representation can be quantified as the level of \textit{overlap} between the actual sample distribution $P_{\mathcal{S}}$ and the expected sample representation $P_{\mathcal{T}}$. To achieve this, we propose an overlap index $V(P_X, P_Y)$ (\ref{eq:cont_ind}), which is a measure of whether the supports of two distributions are similar. With set difference $\Delta$ and 1-d Lebesgue measure (length) of a set $\lambda$, $V$ is essentially the Steinhaus distance \cite{DBLP:conf/cvpr/GardnerKDS14} with an added term $I$ to make $-1 < V < 0$ indicate containment of $P_Y$ within $P_X$. When not contained, for some positive likelihood in both distributions, as illustrated in Figure \ref{fig:ovl_ind}, $V=0$ when they exactly overlap, $V=1$ when they do not overlap, and $0 < V < 1$ when the overlap is partial. Indices $V^j(P_\mathcal{T})$ (\ref{eq:mdist}) quantitatively measure the level of overlap between true and expected distributions for each label. Complementing the visual explanation, overlap indices $0.4 < V^j(P_\mathcal{T}) < 1$ seen in Figure \ref{fig:os_cs_ann}, indicate that there is only slight partial overlap between expectations and reality, confirming notable sample selection bias and, therefore, significant sample under-representation. \looseness=-1
\begin{gather}
    R_X = \{x \in \mathbb{R} : P_X(x) > 0\} \label{eq:supp}\\
    V(P_X, P_Y) = I~\frac{\lambda(R_X ~\Delta ~R_Y)}{{\lambda(R_X ~\cup ~R_Y)}}, 
    ~~I = \left\{ \begin{array}{cc} 
                -1 & \hspace{2mm} R_Y \subset R_X \\
                +1 & \hspace{2mm} otherwise
                \end{array} \right. 
                \label{eq:cont_ind} \\
    V^j(P) = V(P_\mathcal{S}(Y^j~|Y^1),P(Y^j~|Y^1)), ~~j=2...6 \label{eq:mdist} 
\end{gather}

It is therefore clear that, given the expected representation and actual distribution of labels in the dataset, it is possible to comprehensibly explain sample under-representation both visually and quantitatively. However, the overlap index, which eschews mass and uses only support, is an incomplete measure of sample selection bias, the pros and cons of which is discussed in Section \ref{sec:discussion}.

\section{Explaining sample representation using simulation}
The dataset $\mathcal{S}$ contains information $X_i$ in the image domain, while lacking information $Y_i$ in the annotations domain. Expectations, on the contrary, are expressed using annotations $\hat{Y}_i\sim P_\mathcal{T}(Y)$, but lacks images. It is this gap in information that prevents estimation of sample under-representation by direct comparison. There are two possible ways to bridge this gap, one of which is the labeling scheme $Y_i = L(X_i)$ introduced earlier. Another way could be to generate images $\hat{X}_i = G(\hat{Y}_i)$, which is essentially \textit{parametric simulation}. In this case of circles and squares, it is possible to use a graphics package\footnote{We use OpenCV -- https://opencv.org/} to draw shapes using size, position, and brightness labels as parameters. We, in fact, choose this simple dataset because both labeling and simulation of samples are easy, helping illustrate both ways of bridging the gap and cross-checking the plausibility of estimating sample representation. In many practical cases, however, the right method to bridge the gap is difficult to judge since the relative expense is domain and problem specific. Addressing those numerous instances where unlabeled data is available and labeling is expensive, we now show that it is possible to bridge the gap using simulation. This is done using a two-step process, described below, of (i) detecting outlier annotations and (ii) estimating marginal sample representation.\looseness=-1

\subsubsection{Step 1 - Detecting outlier annotations}
\begin{figure}[h]
  \begin{minipage}[t]{0.5\textwidth}
    \centering
    \resizebox{6.0cm}{!}{
    \begin{tikzpicture}[
          line/.style={thick},
          network/.style={draw, trapezium, line width=1.2pt},
          arrow/.style={thick, ->}
        ]
        \node[network, shape border rotate=90] (simulator) {Simulator $(G)$};
        \node[network, shape border rotate=270, my right of=simulator, align=center] (classifier) {Classifier $(F)$};
        
        \node[auto, left= 0.6 of simulator] (yi) {$\hat{Y}_i \sim P_\mathcal{T}(Y)$};
        \node[auto, align=left, right= 0.4 of classifier] (max) {\(max\)};
        \node[auto, align=left, right= 0.4 of max] (S) {$S_i$};
        \node[auto, below= 1.0 cm of classifier] (T) {$T$};
        
        \draw[->] (yi) edge (simulator)
                  (simulator) edge node[auto] (xi) {$\hat{X}_i$} (classifier)
                  (classifier) edge (max)
                  (max) edge (S)
                  (T) edge (classifier)
        ;

        
    \end{tikzpicture}
    }
    \caption{Detecting outlier annotations}
    \label{fig:nov_det}
  \end{minipage}
  \begin{minipage}[t]{0.5\textwidth}
        \centering
        \includegraphics[width=1.0\linewidth]{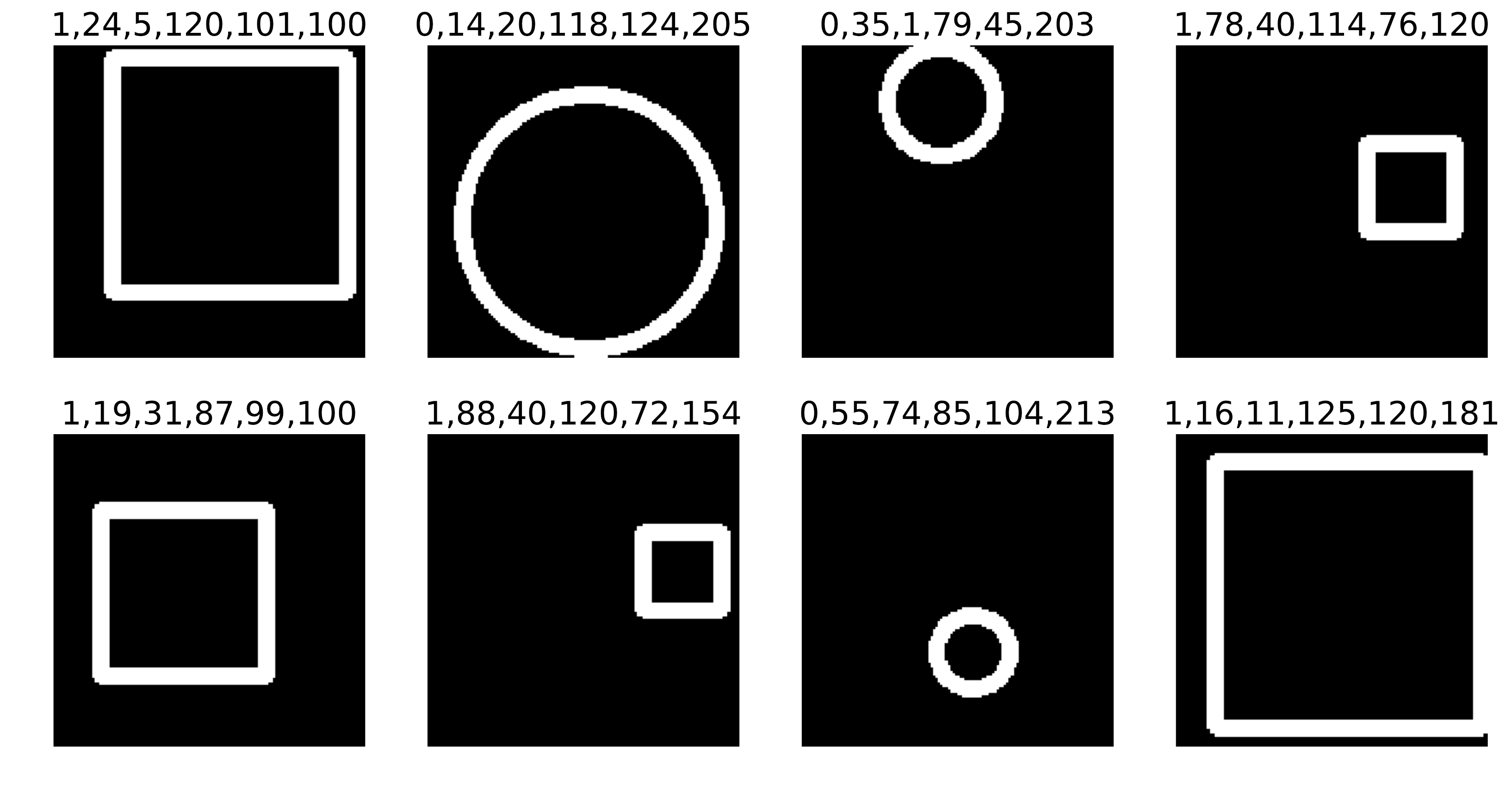}
        \caption{Samples $\hat{X}_i$ from the test set $\mathcal{T}$}
        \label{fig:sd_samples}
  \end{minipage}
  \setlength{\belowcaptionskip}{-20pt}
\end{figure}

To a dataset that mainly contains large, centered shapes, can simulated small off-centered shapes appear as outliers? In order to explore this simple notion, we pose the following outlier hypothesis - \textit{a test annotation $\hat{Y}_i$, that is unlikely to be observed in $\mathcal{S}$, maps to a simulated test sample $\hat{X}_i = G(\hat{Y}_i)$, that appears as an outlier to $\mathcal{S}$}. Bridging the gap by simulating shape images that follow specified expectations $P_\mathcal{T}$, the problem of detecting sample selection bias turns into one of detecting outlier images. The hypothesis is realized by an outlier detector $E_\mathcal{S}$ (Figure \ref{fig:nov_det}) that samples test annotations from $P_\mathcal{T}$ and maps them into images using a simulator, creating a test set $\mathcal{T} = \{(\hat{X}_i, \hat{Y}_i)\},~i=1...M$ (examples in Figure \ref{fig:sd_samples}). Following \cite{DBLP:journals/corr/HendrycksG16c}, the subsequent assessment of whether under-represented simulated images appear as outliers to $\mathcal{S}$ is done using the predictive certainty of a shape label classifier $F(X) =P_{\mathcal{S}}(Y^1 | X;\theta)$, trained on the dataset $\mathcal{S}$.  The complete detector of outlier annotations $E_\mathcal{S}$ is formally described below in (\ref{eq:nov_det}), where $F_k$ is the logit score for the $k^{th}$ shape and $T$ is the temperature parameter which, as shown later, eases the detection process. With $F$ using a softmax output layer, we use maximum softmax score as the measure of certainty. Put simply, with sets of outlier and familiar annotations (\ref{eq:idod}), the outlier hypothesis asserts that a good detector $E_\mathcal{S}$ assigns low scores $S_i$ for outlier annotations $\hat{Y}^-$ and high scores for familiar ones $\hat{Y}^+$. \looseness=-1
\begin{gather}
S_i = E_\mathcal{S}(\hat{Y}_i, F, T) = \mathop{max}_{k \in K}~\frac{exp(F_k(G(\hat{Y}_i))/T)}{\sum\limits_{k \in K} exp(F_k(G(\hat{Y}_i))/T)}, \hat{Y}_i \sim P_\mathcal{T}(Y), K = \{0,1\} \label{eq:nov_det} \\
\hat{Y}^- = \{{\hat{Y}_i : P_\mathcal{S}(\hat{Y}_i) = 0}\},
    \hspace{10mm} \hat{Y}^+ = \{{\hat{Y}_i : P_\mathcal{S}(\hat{Y}_i) > 0}\} \label{eq:idod}
\end{gather}
\vspace{-20pt}

\vspace{-10pt}

\begin{figure}[h]
    \begin{subfigure}[t]{0.5\textwidth}
        \centering    
        \includegraphics[width=1.0\linewidth]{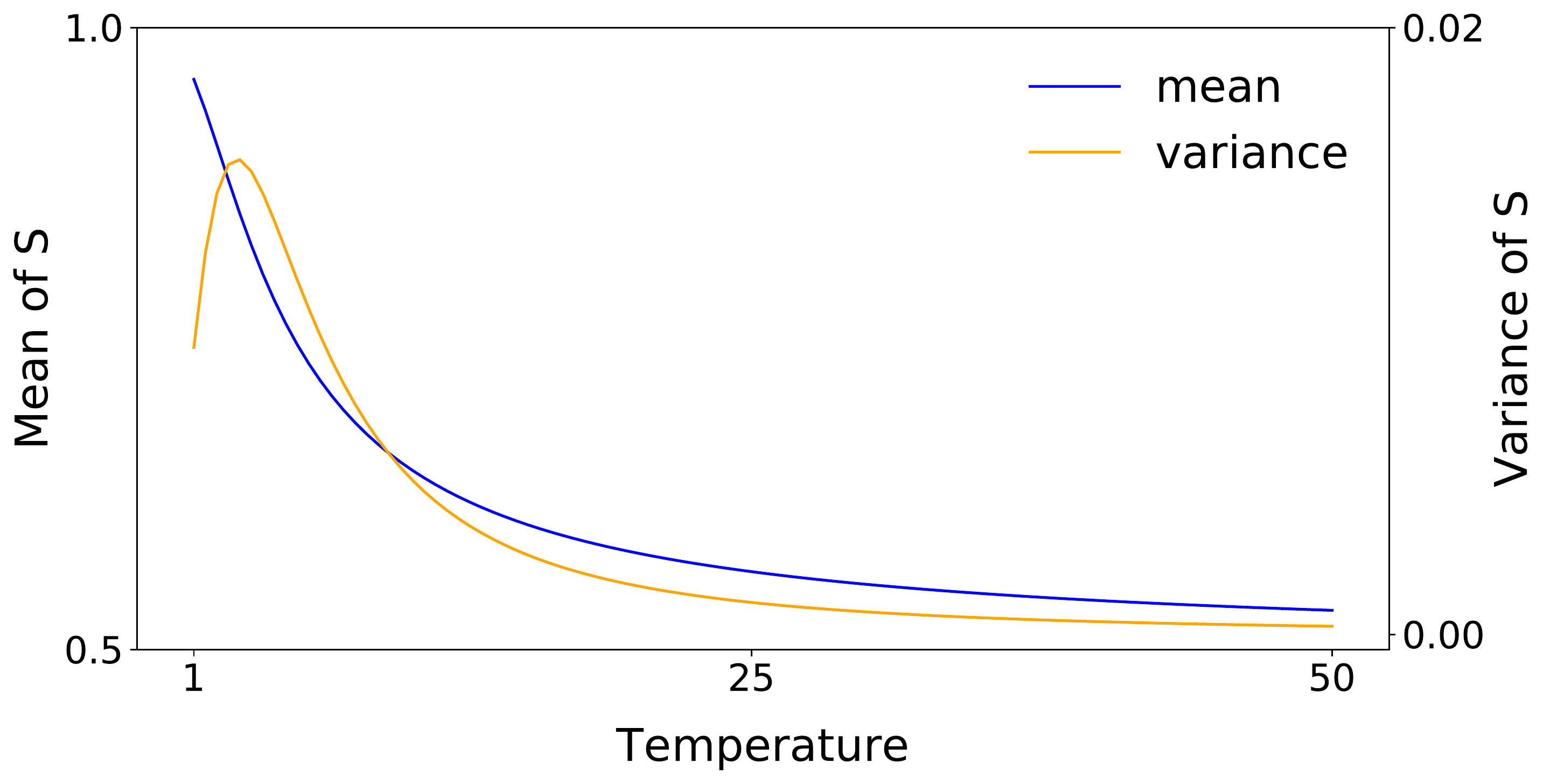}
        \caption{Effect of temperature scaling on the distribution of uncertainty scores $F$=VGG13}
        \label{fig:soft_vgg11}
    \end{subfigure}
    \begin{subfigure}[t]{0.5\textwidth}
        \centering    
        \includegraphics[width=1.0\linewidth]{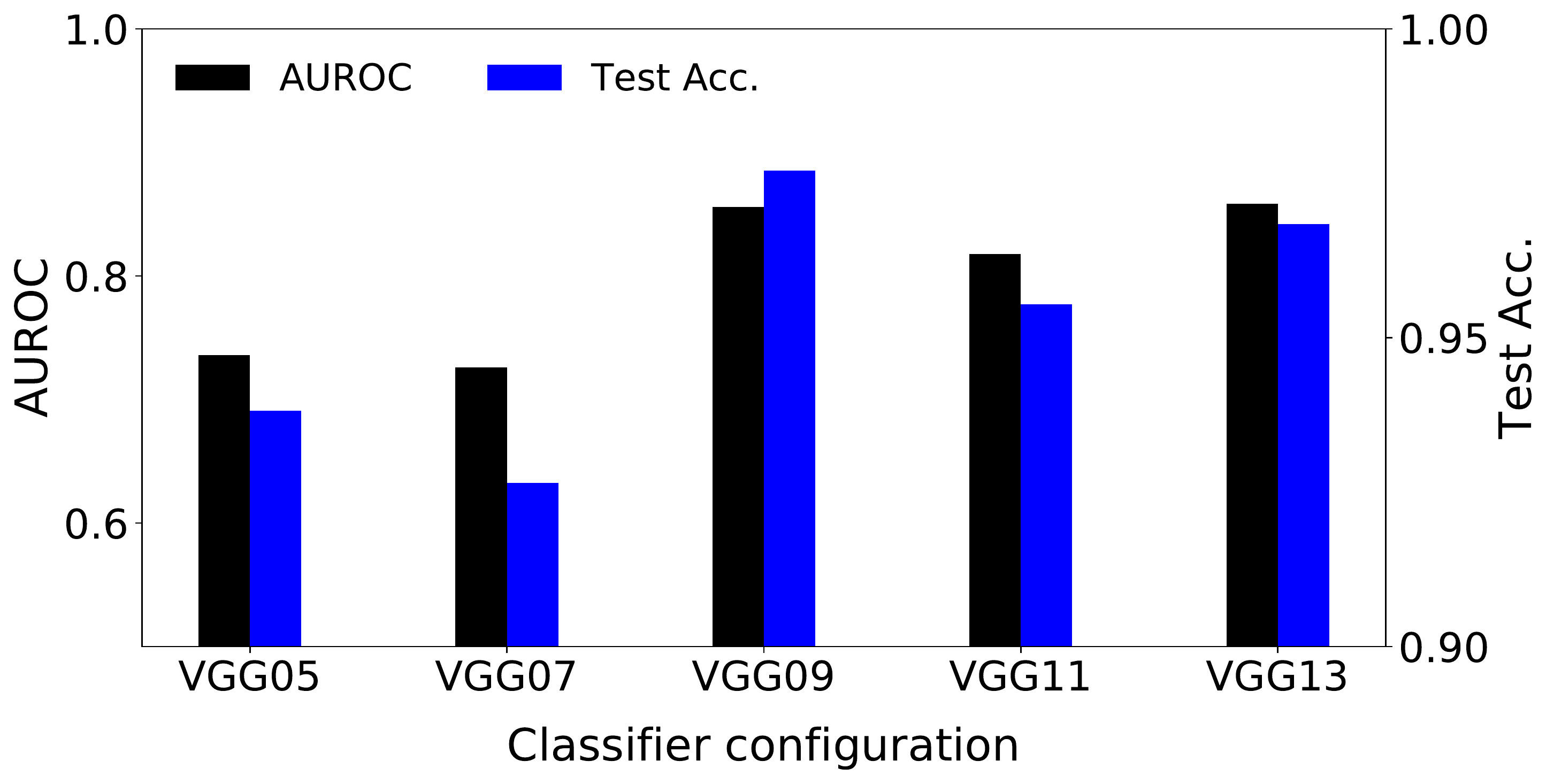}
        \caption{AUROC for detecting outlier annotations per classifier at $T=T^*$ and $S^T=0.7$\looseness=-1}
        \label{fig:tm_roc}
    \end{subfigure}
    \setlength{\belowcaptionskip}{-20pt}
    \caption{Testing the novelty hypothesis}
\end{figure}

To test the outlier hypothesis, four variants of the classifier $F$, all of which follow the VGG architecture \cite{DBLP:journals/corr/SimonyanZ14a}, are used. Classifiers mainly differ in the number of layers, with VGG05 ($5$ layers) and VGG13 ($13$ layers) being the shallowest and deepest respectively. Each $F$ is trained\footnote{Each classifier trains within 10 -- 15 minutes on an NVidia GTX 1080 Ti GPU} for $5$ epochs on $\mathcal{S}$ with 50k samples using the Adam optimizer \cite{DBLP:journals/corr/KingmaB14} to achieve validation accuracy (on a separate set of 10k samples) greater than $97\%$. However, \cite{DBLP:conf/icml/GuoPSW17} shows that deep neural nets tend to predict with high confidence, making raw maximum softmax scores poor measures of predictive certainty, and a simple way to mitigate this is temperature scaling, i.e. setting $T>1$, in (\ref{eq:nov_det}). As seen in Figure \ref{fig:soft_vgg11}, scores $S_i$ are tightly clustered at $T=1$ with relatively low variance, which makes it difficult to identify differences in predictive certainty between familiar and outlier annotations. There is, however, a range of temperatures at which scores are better spread and can exaggerate these differences. While a temperature that maximizes the variance of the score distribution seems appropriate, as seen in Figure \ref{fig:soft_vgg11}, scaling also reduces its mean. Therefore a safeguard may be necessary to prevent the mean certainty score from reducing to a level that questions the confidence of predictions. These twin requirements can be achieved by the search objective (\ref{eq:tm_obj}), which ensures a good spread in scores $S_i$, while keeping its mean close to the chosen safeguard $S^T$. \looseness=-1
\begin{equation}
\label{eq:tm_obj}
\begin{gathered}
T^* =  \mathop{argmin}_{T}~ L^T~-~L^V,
\hspace{3mm}L^T = \left(\mu_S - S^T\right)^2\\
L^V = \frac{\sum_{i=1}^M\left(E_{\mathcal{S}}(\hat{Y}_i, F, T) - \mu_{S}\right)^2}{M},
\hspace{3mm}\mu_S = \frac{\sum_{i=1}^M E_{\mathcal{S}}(\hat{Y}_i, F, T)}{M}
\end{gathered}
\end{equation}
Upon temperature scaling with $T^*$, the effectiveness of the detector $E_\mathcal{S}$ in separating outlier annotations $\hat{Y}^-$ from familiar ones $\hat{Y}^+$ can be measured using the Area Under Receiver Operating Characteristic (AUROC). This is shown for each $F$, averaged over $5$ separate training runs, in Figure \ref{fig:tm_roc}. Based on an informal grading scheme for classifiers using AUROC score suggested in \cite{DBLP:journals/corr/HendrycksG16c}\footnote{Quality of classification based on AUROC score - 0.9—1: Excellent, 0.8—0.9: Good, 0.7—0.8: Fair, 0.6—0.7: Poor, 0.5—0.6: Fail}, detectors using VGG05 and VGG07 receive a {\lq{fair}\rq} grade in identifying outlier annotations, while the deeper networks get {\lq{good}\rq} grades. The best outlier detectors, with AUROC $\approx 0.85$, are those with $F$ as VGG09 and VGG13. These results clearly endorse the viability of the outlier hypothesis that simulated images that are under-represented in $\mathcal{S}$, in terms of specified design concerns, appear as outliers to the right classifier trained on $\mathcal{S}$. While $P_\mathcal{S}$, derived from labeling, is used as a benchmark to test the outlier hypothesis, it is important to observe that (i) classifiers that are good at outlier detection are, as seen in Figure \ref{fig:tm_roc}, those that have the highest accuracy in predicting shape labels on the test set $\mathcal{T}$, and (ii) the temperature $T^*$, at which the classifiers become good outlier detectors, depends only upon the statistical properties of scores $S_i$. Together, these observations mean that a good detector of under-represented annotations can be assembled using only simulation, without any need for labeling. 

\subsubsection{Step 2 - Estimating marginal sample representation}
As presented in Section \ref{sec:exp_bias_ann}, we seek to comprehensibly explain sample representation in the dataset $\mathcal{S}$ of geometric shapes on the basis of intuitive design concerns like size, position, and brightness. However, the detector $E_\mathcal{S}$ can only assess whether a single combined 6-d test annotation is an outlier. To assess, for example, the diversity of shape sizes in the dataset, independent of position, we turn to techniques of input attribution. Given the detector $E_\mathcal{S}$, attribution techniques estimate the contribution of each input label $\hat{Y}^j_i$ to its outlier score $S_i$. Among proposed methods for input attribution \cite{46831}, one promising framework is Shapley Additive Explanations (SHAP)\cite{DBLP:conf/nips/LundbergL17}. Using principles of cooperative game theory, SHAP estimates \textit{marginal influence} $\phi^j_i$ (\ref{eq:shap}), which indicates how label $\hat{Y}^j_i$ independently influences the uncertainty score $S_i$.\looseness=-1
\begin{gather}
    \label{eq:shap}
    S_i = E_\mathcal{S}(\hat{Y}_i, F, T) = \phi^{0} + \sum_{j=2}^6 \phi^j_i
\end{gather}
In satisfying an additive property, SHAP values are also semantically intuitive, with negative, positive, and zero values of $\phi^j_i$ respectively indicating negative, positive, and neutral influence of label $\hat{Y}^j_i$ on the score $S_i$.  The  outlier hypothesis verified earlier implies that outlier (familiar) annotations tend to have a lower (higher) certainty score $S_i$. Therefore SHAP value $\phi^j_i > 0$, which indicates that the individual label value $\hat{Y}^j_i$ tends to improve $S_i$, becomes an indicator of that label being represented in $\mathcal{S}$. Through a campaign directed by the test set $\mathcal{T}$, which systematically covers the specified range of scenarios $P_\mathcal{T}$, non-negative SHAP values identify sample representation in the dataset $\mathcal{S}$ in terms of each individual label. This can be seen in Figure \ref{fig:stcal_shap_vgg13}, where label values with a high incidence of non-negative SHAP values (marked black) are likely to be represented in $\mathcal{S}$. This directly allows estimating the likelihood of test label $Y^j=l, ~ Y^j \sim P_\mathcal{T}$ being represented in the set $\mathcal{S}$ as the proportion of test labels $\hat{Y}^j_i$, in a sufficiently small interval $\delta$ around $l$, whose SHAP values are non-negative.
\begin{equation}
\begin{gathered}
    P^+_{\mathcal{T}}(Y^j=l~|Y^1=k) = \frac{|\{\hat{Y}^j_i : \phi_i^j \geq 0,~\hat{Y}_i^j \in Y^l\}|}{|\{\hat{Y}^j_i : \phi_i^j \geq 0\}|},~j=2...6,~\hat{Y}_i \in \hat{Y} \\
    Y^l =  \{l-\delta, l+\delta\},
    \hspace{10mm} \hat{Y} = \{\hat{Y}_i : \hat{Y}^1_i = k\}, ~k \in K
\end{gathered}    
\end{equation}

\subsubsection{Assessing the explanation}
\begin{figure}
    \begin{subfigure}[t]{0.5\textwidth}
        \centering    
        \includegraphics[width=1.0\linewidth]{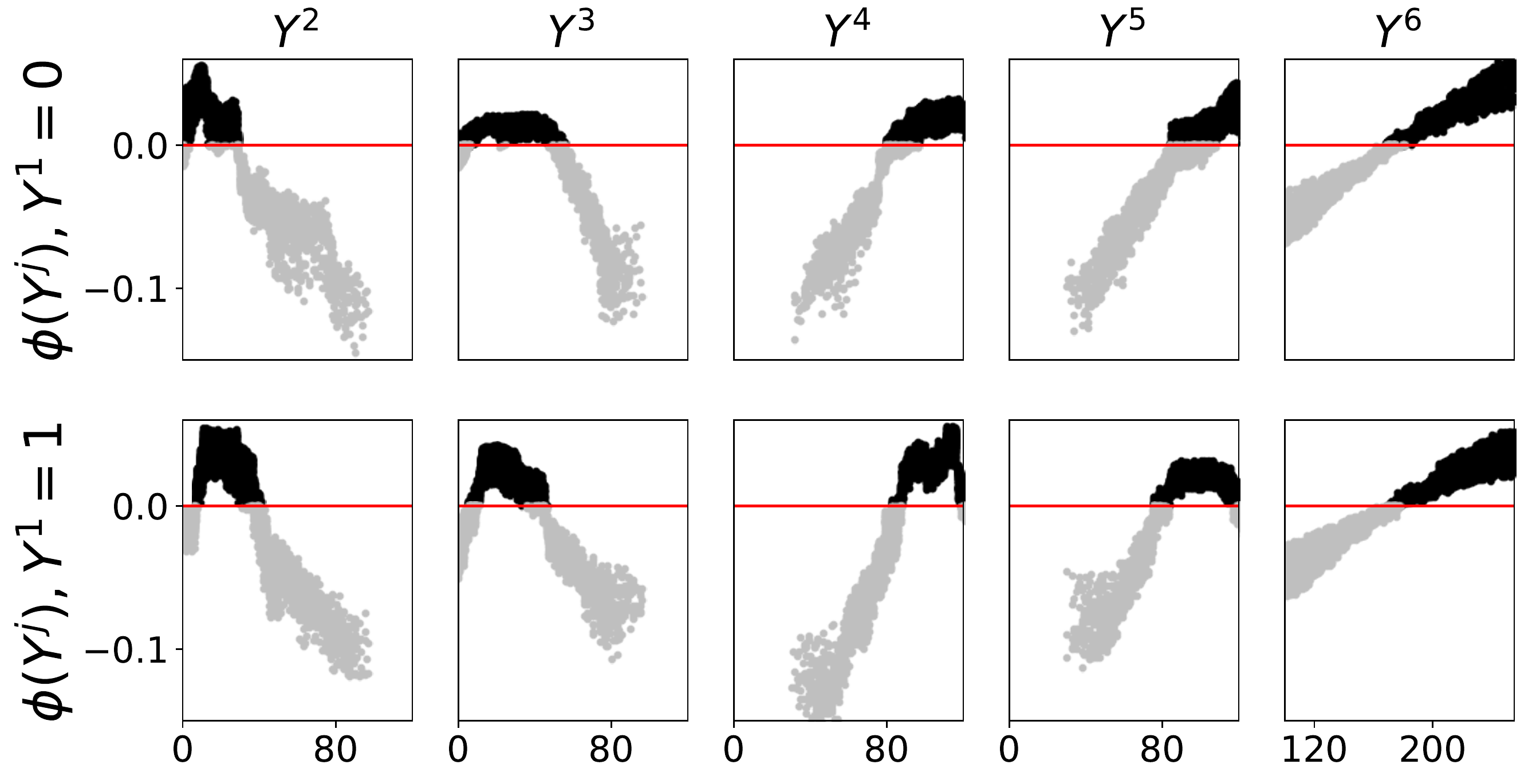}
        \caption{Sample-representation from SHAP scores}
        \label{fig:stcal_shap_vgg13}
    \end{subfigure}
    \begin{subfigure}[t]{0.5\textwidth}
        \centering    
        \includegraphics[width=1.0\linewidth]{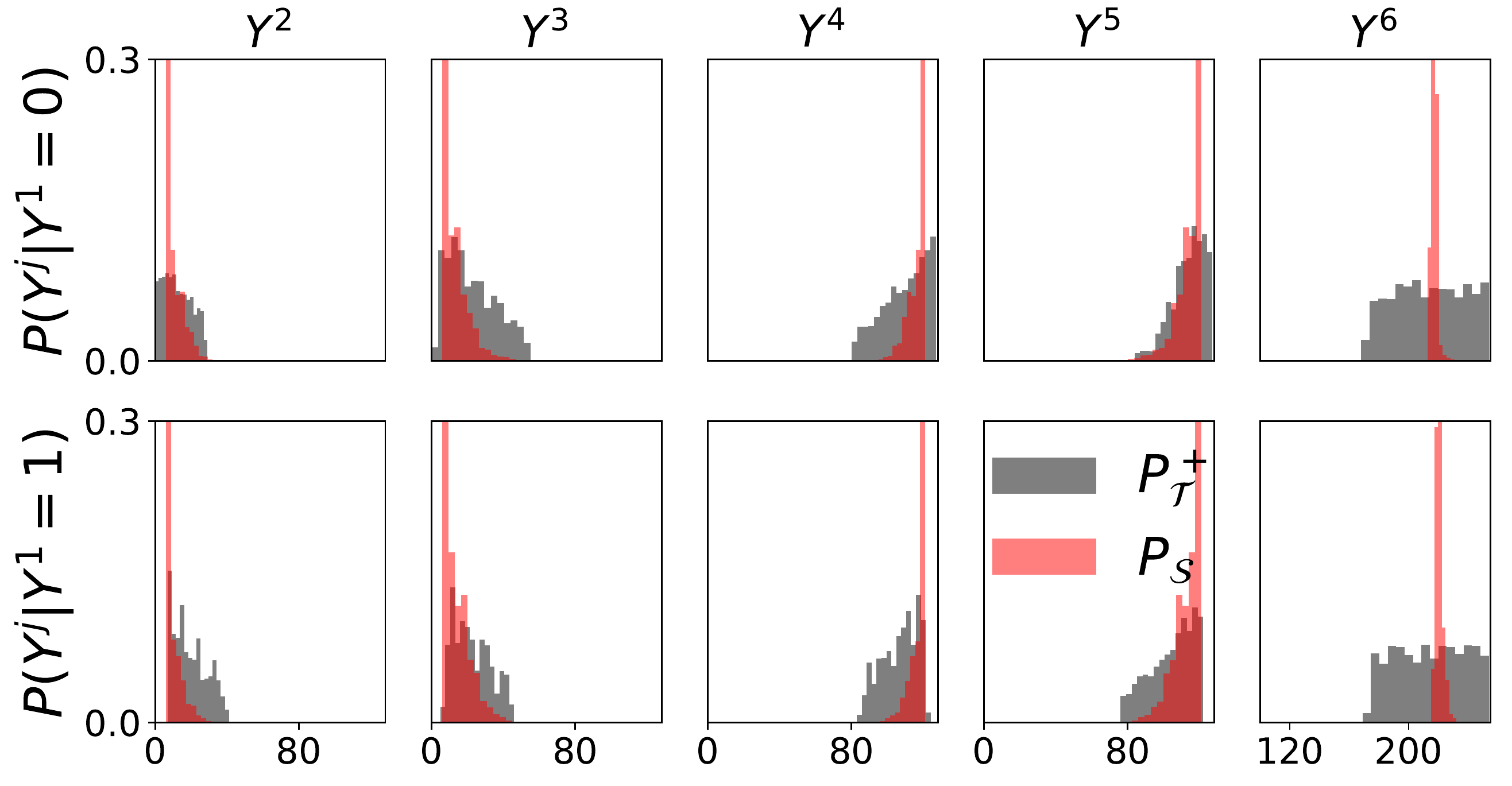}
        \caption{Marginal sample representation}
        \label{fig:stcal_marg_vgg13}
    \end{subfigure}
    \setlength{\belowcaptionskip}{-10pt}
    \caption{Explaining sample representation using simulation ($F$=VGG13, $T=T^*$, $S^T=0.7$)}
\end{figure}

By expressing expected diversity $P_\mathcal{T}$ in terms of specified design concerns, the two-step process, using a simulated test set, identifies sample representation in each concern using non-negative influence on predictive certainty. From the original broadly spread expectations $P_\mathcal{T}$ (Figure \ref{fig:os_cs_ann}), the process correctly eliminates a significant amount of outliers in each label dimension, producing $P^+_\mathcal{T}$ (Figure \ref{fig:stcal_marg_vgg13}). $P^+_\mathcal{T}$ shows label values likely to be observed in the dataset $\mathcal{S}$ and has a roughly similar spread as the actual distribution $P_\mathcal{S}$. Also, using a test set with $M$=10k samples, the process estimates sample representation in a much larger dataset with $N$=50k samples.\looseness=-1

Introduced originally in Section \ref{sec:exp_bias_ann} to quantify bias between expected and actual distributions of annotations, the overlap index $V$ is also suitable for measuring similarity between $P^+_\mathcal{T}$ and $P_\mathcal{S}$. This helps quantify the effectiveness of estimating sample representation using simulation. The visual observation that $P^+_T$ is a better estimate of true sample distribution, compared to the broad range of expectations $P_\mathcal{T}$, is confirmed by better a mean overlap score  $V^j(P^+_\mathcal{T})$ (see Table \ref{tab:quant_bias}), over all labels and shapes, compared to mean $V^j(P_\mathcal{T})$. While this holds true for both classifier instances shown in the table, the detector using $F$=VGG13 at $T=T^*$, which has the best AUROC score in detecting outliers, produces the closest estimate with a mean overlap score of $0.27$. VGG05, with poorer AUROC, has a weaker average overlap score of $0.39$. The close correlation between AUROC and $V$ further confirms the plausibility of estimating marginal sample representation using SHAP scores. This shows that, while facing an expensive labeling process, with the right means of parametric simulation, one can conduct a campaign from a low-dimensional space of specified design concerns to estimate sample representation in a given dataset and comprehensibly explain sample selection bias. \looseness=-1
\vspace{-15pt}

\section{Discussion}
\label{sec:discussion}
\subsubsection{Under-representation and outlier detection}
\begin{figure}
  \begin{minipage}[b]{0.5\textwidth}
  \setlength{\extrarowheight}{5pt}
    \centering
    \scalebox{0.55}{
    \begin{tabular}{|l|l|c|c|c|c|c|c|c|}
    \hline
    \multirow{2}{*}{~~~$\boldsymbol{T}$~~~} & \multirow{2}{*}{~~~$\boldsymbol{P}$~~~} & \multirow{2}{*}{~$\boldsymbol{Y^1}$~} & \multicolumn{5}{c|}{$\boldsymbol{V^j(P)}$} &
    \multirow{2}{*}{\shortstack[c]{\textbf{Mean}\\ $\boldsymbol{V^j(P)}$}}\\ \cline{4-8}
    & & & ~$\boldsymbol{j}$=\textbf{2}~ & ~~~\textbf{3}~~~ & ~~~\textbf{4}~~~ & ~~~\textbf{5}~~~ & ~~~\textbf{6}~~~ &\\ \hline
    
    \multirow{2}{*}{~-~} & \multirow{2}{*}{~$P_\mathcal{T}$} & 0 & 
    0.60 & 0.47 & 0.59 & 0.47 & 0.75 &
    \multirow{2}{*}{0.57}\\ \cline{3-8}
    & & 1 &
    0.57 & 0.45 & 0.57 & 0.45 & 0.76
    & \\ \hline
    
    \multirow{4}{*}{~\shortstack[l]{$T^*$\\ $S^T=0.7$}~} & \multirow{2}{*}{~\shortstack[l]{$P^+_\mathcal{T}$\\ $F=$VGG13}~} & 0 & 
    0.49 & 0.14 & 0.17 & 0.35 & 0.55 &
    \multirow{2}{*}{0.27}\\ \cline{3-8}
    & & 1 &
    -0.19 & 0.26 & 0.16 & 0.17 & 0.56
    & \\\cline{2-9}
    
    & \multirow{2}{*}{~\shortstack[l]{$P^+_\mathcal{T}$\\ $F=$VGG05}~} & 0 & 
    0.47 & 0.33 & 0.29 & 0.44 & 0.60 &
    \multirow{2}{*}{0.39}\\ \cline{3-8}
    & & 1 &
    0.31 & 0.34 & 0.27 & 0.25 & 0.56
    & \\ \hline
    
    \multirow{4}{*}{~$1$~} & \multirow{2}{*}{~\shortstack[l]{$P^+_\mathcal{T}$\\ $F=$VGG13}~} & 0 &
    0.49 & 0.30 & 0.40 & 0.15 & 0.69 &
    \multirow{2}{*}{0.36}\\ \cline{3-8}
    & & 1 &
    0.30 & 0.28 & 0.21 & 0.13 & 0.69
    & \\\cline{2-9}
    
    & \multirow{2}{*}{~\shortstack[l]{$P^+_\mathcal{T}$\\ $F=$VGG05}~} & 0 &
    0.22 & 0.14 & 0.54 & 0.47 & 0.65 &
    \multirow{2}{*}{0.43}\\ \cline{3-8}
    & & 1 &
    0.57 & 0.29 & 0.56 & 0.15 & 0.70
    & \\ \hline
    \end{tabular}
    }
    \setlength{\belowcaptionskip}{-8pt}
    \captionof{table}{Quantitative bias estimation}
    \label{tab:quant_bias}
    \end{minipage}
    \hfill    
  \begin{minipage}[b]{0.5\textwidth}
    \centering
    \includegraphics[width=\linewidth]{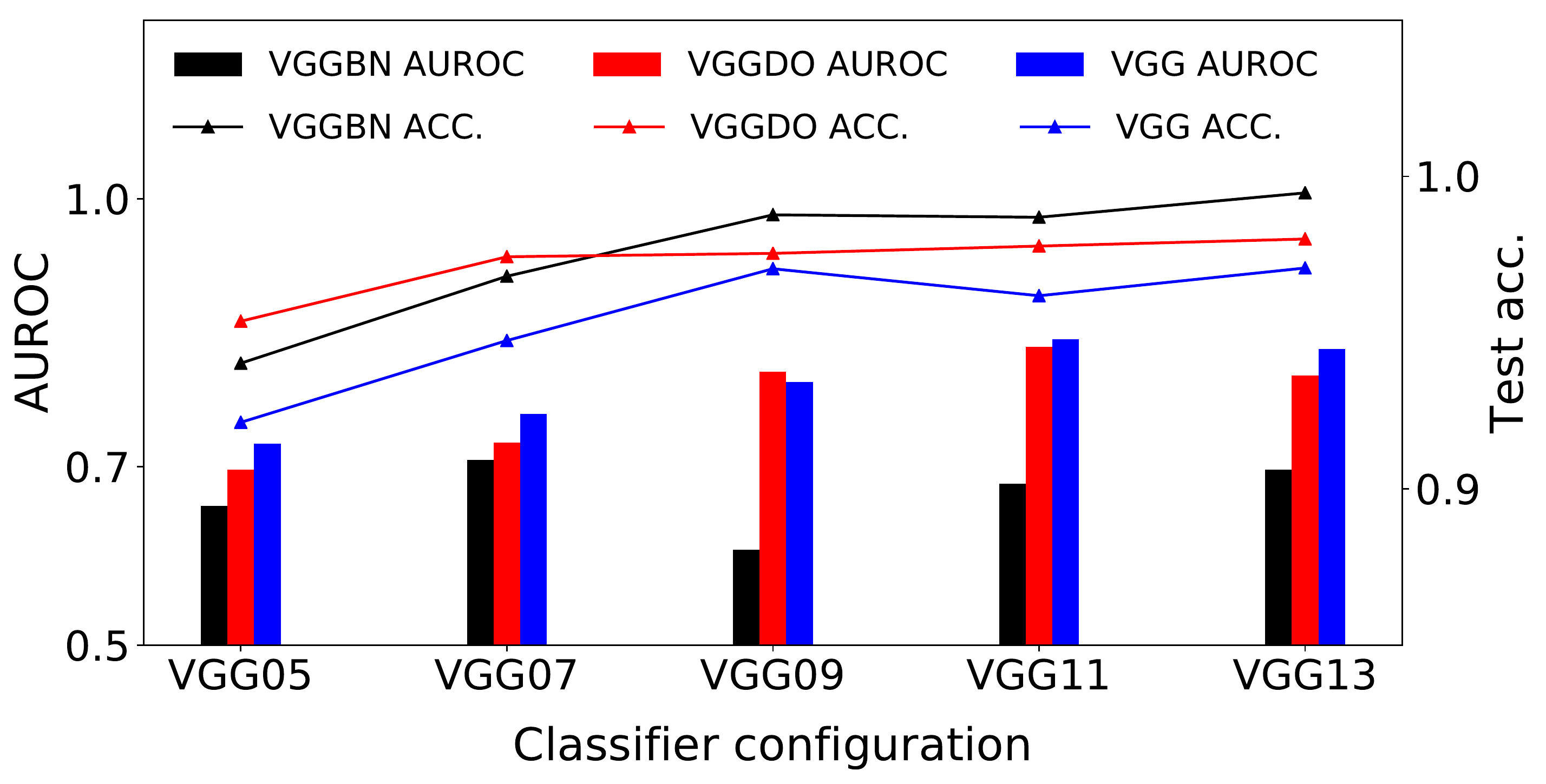}
    \setlength{\belowcaptionskip}{-6pt}
    \captionof{figure}{Effect of regularization on AUROC}
    \label{fig:reg_roc}
  \end{minipage}
  \end{figure}

A good outlier detector $E_\mathcal{S}$ of under-represented samples must blur the distinction between simulated and real images while emphasizing the distinction between over and under-represented images. Figure \ref{fig:tm_roc} shows both conditions are jointly achievable, with classifiers that have a high test set accuracy, and therefore generalize well, also having better AUROC scores in detecting representation. However, as seen in Figure \ref{fig:reg_roc}, using regularization measures like batch normalization layers after each convolutional block, while improving test accuracy, reduces AUROC scores for all classifier instances. This is probably because it tends to blur \cite{DBLP:conf/iclr/LiWS0H17} both forms of distinction. The figure also shows that dropout increases the test accuracy without any major effect on AUROC scores, giving no special domain separation advantage in detecting under-representation. Among the classifier configurations investigated here, vanilla VGG, with the strongest correlation between AUROC and test set accuracy, is observed to best addresses both forms of domain distinction. \looseness=-1

\subsubsection{The importance of effective simulation}
It is crucial to note that high test accuracy reflects the combined effect of plausible simulation and good generalization. It is equally essential, therefore, that the simulator produces samples that are plausibly real. Ensuring effective simulation, while supporting a variety of parameters, is undoubtedly a challenge for realistic datasets with richer content. As noted earlier, while this is domain and problem dependent, for images at least, rapid advancements in the quality and range of graphics tools (\cite{DBLP:journals/cgf/ChaoBLMWLD20},\cite{DBLP:journals/vc/ErsotelosD08}), potentially makes effective simulation plausible. However, with notable progress in techniques that automate parts of the labeling process \cite{DBLP:journals/pr/ChengZ0TL18}, it is also important to assess whether labeling is cheaper for the dataset in concern.

\subsubsection{Improving estimation of representation}
Figure \ref{fig:stcal_marg_vgg13} shows that while the estimated sample representation $P^+_\mathcal{T}$ comes close, it does not overlap perfectly with the true label distribution $P_\mathcal{S}$. As quantified in Table \ref{tab:quant_bias}, even the best detector ($F$=VGG13 at $T=T^*$) has a mean overlap index of $0.27$ indicating relatively close, but only partial, overlap on average. At the individual label level, index values show varying accuracy in support-matching. The representation of pixel brightness $0.5 < V^6(P_\mathcal{T}) < 0.8$ is consistently underestimated, while those of bounding box coordinates are better estimated. It is however clear from Table \ref{tab:quant_bias} that temperature scaling ($T=T^*$ vs $1$) and deeper classifiers ($F=$VGG13 vs VGG05) improve estimation, indicating that more sophisticated techniques of predictive outlier detection, like methods in \cite{DBLP:conf/bmvc/ShafaeiSL19}, can improve estimation.\looseness=-1

\subsubsection{Balancing detail in specifying expectations} 
The level of detail specified in the expectations $P_{\mathcal{T}}$ plays a key role in deciding the cost and benefit of explaining sample representation. An overly detailed breakdown of design factors involves significant engineering effort, degrades interpretability, and overlooks the remarkable benefits of generalization offered by deep learning. But well-balanced expectations can provide valuable insight into training data. Take an application like self-driving vehicles, where engineers actively seek a certain level of understanding of operational scenarios \cite{gyllenhammar:hal-02456077} to ensure safe operation. Such understanding can be exploited to systematically explain, analyze, and manage the data used to train models deployed in the system, thereby improving overall confidence in its dependability. While balancing details in the specification may not always be easy, one advantage of this method is that it is semi-supervised. Annotations included in the analysis impacts only the simulated test set $\mathcal{T}$ and has no effect on the actual dataset $\mathcal{S}$.\looseness=-1

\subsubsection{Extension to other domains}
This method of explanation can conceivably be extended to a problem in another domain if (i) operational scenarios can be reasonably broken down and (ii) model-based parametric simulators that can generate data for this domain are available. For example, this method can use a simulator of vulnerable road user trajectories \cite{Helbing_1995} to examine a sparsely labeled dataset of trajectories (e.g. \cite{DBLP:conf/iccv/RasouliKKT19}) and check whether it adequately represents trajectories of risk groups like elder pedestrians, electric bikes, etc.

\section{Related work}
\subsubsection{Sample selection bias}
Sample selection bias has been addressed in existing literature from the perspective of domain adaptation \cite{DBLP:journals/corr/abs-1812-11806}. Previous methods to mitigate sample selection bias have mainly attempted to modify the training procedures or the model itself to yield classifiers that work well on the test distribution. Methods such as importance re-weighting \cite{DBLP:phd/hal/Tran17b}, minimax optimization \cite{DBLP:conf/nips/LiuZ14}, kernel density estimation \cite{DBLP:conf/nips/DudikSP05} and model averaging \cite{DBLP:conf/sdm/FanD07} all fall in this category. While these methods can yield classifiers that are able to generalize, the accuracy can suffer when the two distributions differ greatly in the overlap of their support or in the distribution of their mass. Our immediate goal, on the other hand, does not seek to obtain a classifier that generalizes, but instead we seek to obtain a high level \textit{understanding} of the deficiencies of our training data and where the bias stems from. This goal does not necessarily require a full specification of $P(Y)$, instead we work with the weak proxy of $P_{\mathcal{T}}(Y)$ which attempts to match $P(Y)$ only through the support. However, by eschewing mass-modeling,  we gain a few advantages, one of which is the reduced effort in defining expectations. More importantly, since several existing methods for correcting sample selection bias work only if the support of $P_{\mathcal{T}}$ is included in that of $P_{\mathcal{S}}$ and our method of explanation tests precisely for this condition. Overlap indices $V^j(P_{\mathcal{T}}) \leq 0$ guarantees that the support of the biased distribution includes that of the expectations and correction measures like importance re-weighting are applicable. If $ 0 < V^j(P_{\mathcal{T}}) \leq 1$, expanding the diversity of data collection is unavoidable. Thus seeking to understand and explain the data set can allow for an improved understanding of the validity for methods that directly impacts the generalization performance. \looseness=-1

\subsubsection{Understanding sample representation}Besides clustering approaches \cite{DBLP:conf/icdm/ChenCHCCSD16} and feature projection methods such as t-SNE \cite{vanDerMaaten2008}, previous research into providing a high level understanding of the training set has, for example, applied tree-based methods to detect regions of low point density in the input space \cite{DBLP:conf/iccps/GuE19}. High-dimensional explanations in the input space, however, adversely affects interpretation, and ways to extend these methods to yield explanations using an interpretable low-dimensional space of annotations are not immediately clear. \looseness=-1

\subsubsection{Bias estimation using simulation} Closer to our purpose are the methods \cite{DBLP:conf/nips/McDuffMSK19} and \cite{DBLP:journals/corr/abs-1810-00471} which detect inherent biases in a trained model using parametric simulation and Bayesian optimization. While their goal is to find input samples where the model is locally weak, our goal is to ensure that a given dataset meets global expectations defined by a test set. This can verify that a system is dependable for all considered scenarios, like \cite{Thorn2018-09-01}, which is a standardized set of tests. However, in reformulating bias detection as outlier detection, our method -- unlike the aforementioned methods -- trades-off the ability to detect unknown unknowns \cite{DBLP:conf/aaai/LakkarajuKCH17} in favor of a faster, global evaluation of bias. Combining our global and their local approaches may, therefore, help ensure better overall dependability. \looseness=-1

\subsubsection{Shapley-based outlier detection}
Previous work using Shapley values for outlier detection, such as \cite{DBLP:conf/cikm/GiurgiuS19} and \cite{DBLP:journals/corr/abs-1903-02407}, focus mainly on providing interpretable explanations for why a data point is considered to be an outlier. It may also be possible to extend their data-space explanations to the annotation-space, like we do, using parametric simulation. However, pixel-wise reconstruction error has well-known drawbacks in capturing structural aspects of data \cite{DBLP:conf/icml/LarsenSLW16}. It is therefore not immediately clear whether their use of auto-encoder reconstruction error is as good at detecting structural under-representation as our technique of using predictive certainty, which is calculated from the feature space of a classifier. \looseness=-1

\section{Conclusions}
With data playing a crucial role in deciding the behavior of trained models, evaluating whether training and validation sets meets design expectations would be a helpful step towards a better understanding of model properties. To aid this evaluation, we demonstrate a method to specify expectations on and evaluate sample representation in a dataset, in a human interpretable form, in terms of annotations. Using parametric simulation to map test annotations into a test set, the method exposes under-representation by measuring the uncertainty of a classifier, trained on the original dataset, in recognizing test set samples. Techniques of input attribution enable further conversion of predictive uncertainty into a comprehensible low-dimensional estimate of sample representation in the dataset. While refinements in estimation are possible, the core quantitative and qualitative methods shown here are valuable aids in understanding a dataset and, consequently, the properties of a model trained using this data.

\bibliographystyle{splncs04}
\bibliography{bibliography}

\end{document}